\documentclass[final]{cvpr}

\usepackage{times}
\usepackage{epsfig}
\usepackage{graphicx}
\usepackage{amsmath}
\usepackage{amssymb}
\usepackage[numbers]{natbib}

\usepackage{amsmath,amsfonts,bm}

\def\eqref#1{equation~\ref{#1}}
\def\1{\bm{1}}

\DeclareMathAlphabet{\mathsfit}{\encodingdefault}{\sfdefault}{m}{sl}
\SetMathAlphabet{\mathsfit}{bold}{\encodingdefault}{\sfdefault}{bx}{n}

\newcommand{\R}{\mathbb{R}}

\usepackage{url}
\usepackage{graphicx}
\usepackage{multirow}
\usepackage{bold-extra}
\usepackage{makecell}

\usepackage[pagebackref=true,breaklinks=true,colorlinks,bookmarks=false]{hyperref}

\newcommand{\inctabcolsep}[2]{\addtolength{\tabcolsep}{#1} #2 \addtolength{\tabcolsep}{-#1}}

\newcommand{\omt}[1]{}
\def\F{\mathcal{F}}

\title{Compositional Models: Multi-Task Learning and Knowledge Transfer with Modular Networks}

\begin{document}

%%%%%%%%% TITLE
\author{
    Andrey Zhmoginov \& Mark Sandler \\
    Google LLC \\
    1600 Amphitheatre Parkway \\
    Mountain View, CA 94043, USA \\
    \texttt{\{azhmogin,sandler\}@google.com} \\
    \and
    Dina Bashkirova \\
    Boston University \\
    Boston, MA 02215, USA \\
    \texttt{dbash@bu.edu}
}
\maketitle

\begin{abstract}
    Conditional computation and modular networks have been recently proposed for multitask learning and other problems as a way to decompose problem solving into multiple reusable computational blocks.
    We propose a new approach for learning modular networks based on the isometric version of ResNet with all residual blocks having the same configuration and the same number of parameters. This architectural choice allows adding, removing and changing the order of residual blocks.
    In our method, the modules can be invoked {\em repeatedly} and allow knowledge transfer to novel tasks by adjusting the order of computation.
    This allows soft weight sharing between tasks with only a small increase in the number of parameters. 
    We show that our method leads to interpretable self-organization of modules in case of multi-task learning, transfer learning and domain adaptation while achieving competitive results on those tasks.
    From practical perspective, our approach allows to: (a) reuse existing modules for learning new task by adjusting the computation order, (b) use it for unsupervised multi-source domain adaptation to illustrate that adaptation to unseen data can be achieved by only manipulating the order of pretrained modules, (c) show how our approach can be used to increase accuracy of existing architectures for image classification tasks such as \textsc{ImageNet}, without any parameter increase, by reusing the same block multiple times. 
\end{abstract}

\section{Introduction}

    Most modern interpretations of neural networks treat layers as transformations that convert lower-level features, such as pixels, into higher-level abstractions.
    Adopting this view, many designs that break up monolithic models into interacting modular components localize them to particular processing stages thus disallowing parameter sharing across layers.
    This restriction, however, can be limiting when we need to continually grow the model, be that for transferring knowledge to a new task or preventing catastrophic forgetting.
    It also prohibits module reuse within a model even though certain parameter-efficient architectures and complex decision making processes may involve recurrent components \cite{kaiser2016gpu,randazzo2020classifying} and feedback loops \cite{herzog20feedback,yan2019feedback,kar2019evidence}.
    
    Here we propose a simple alternative design that represents an entire model as a mixture of modules each of which can contribute to computation at any processing stage.
    As opposed to many other approaches (\cite{zoph2016neural, bengio2016reinforcement, kirsch2018modular, rosenbaum2019dispatched} etc.), we use a soft mixture of modules to obtain a more flexible model and produce a differentiable optimization objective that can be optimized end-to-end and does not involve high-variance estimators.
    Following our approach, the parameters of every block (or layer) of the network is computed as a linear combination of a set of ``template'' block parameters thus representing the entire model as: (a) a databank of template blocks and (b) vectors of ``mixture weights'' that are used to generate weights for every layer.
    This simple design can be utilized for a variety of applications from producing compact networks and training multi-task models capable of re-using individual modules to knowledge transfer and domain adaptation.
    The experimental results show that: (a) when used for multi-task training, our model organizes its modules where tasks share first few layers while specializing closer to the head, while (b) in domain adaptation problems modules instead specialize on processing the image, while sharing later processing stages.
    Moreover, our self-organizing model achieves promising results in multi-task learning and adaptation to new domains.
    
    Our contributions in this paper can be summarized as follows:
    \begin{itemize}
        \item We propose a method for training self-organizing modular networks capable of arranging available layer weights in an arbitrary order and using them repeatedly across layers and tasks. The proposed architecture is simple and intuitive and does not rely on highly unstable and difficult to train RL-based routing networks.
        Our model allows naturally to: selectively share knowledge between different co-trained tasks via shared templates; train new tasks or adapt to new domains by tuning task-specific mixture weights and potentially adding new trainable templates; scale the model by adding or removing templates.
        \item We show that in accordance with our intuition, modules in this network self-organize to either share early processing stages for images from the same domain, or share final layers for similar tasks trained on different image domains.
        \item Demonstrate that our method achieves state-of-art performance in a challenging ImageNet classification task as well as a set of other commonly used classification benchmarks. 
        \item We show that our architecture is general enough to be applied to a number of problems, including single-task and multi-task learning, transfer learning, and domain adaptation. To our knowledge, we are the first to apply modular networks for the unsupervised domain adaptation problem. We provide the analysis on the efficiency of modular networks for DA. 
    \end{itemize}
    
    The rest of the paper is organized as follows: in Section~\ref{sec:prior_work} we go over the related literature and discuss the existing approaches to modular networks and neural architecture search; in Section~\ref{sec:comp_model} we describe the architecture of our conversational model and various aspects of learning the best module sequence for each specific task; we introduce the experimental results in a single-task setting, multi-task learning, continual learning and domain adaptation in Section~\ref{sec:Experiments} and provide a final discussion in Section~\ref{sec:discussion}.
\section{Prior Work}
\label{sec:prior_work}
    Conditional computation routing is even more appealing for the multitask learning application, as it allows to both learn reusable computation blocks and adjust the model to each specific task with minimal network alteration. For instance, \cite{misra2016cross} use an additional set of modules for each task and enforce similarity between the corresponding task-specific modules weights. \cite{rosenbaum2017routing, rosenbaum2019dispatched} used reinforcement learning to perform task- and data-specific routing; \cite{sun2019adashare} train both the model weights and the task-specific policy that determines which layers should be shared for a given task; \cite{ma2019snr, maziarz2019gumbel} introduced a multi-task model where each layer's output is computed as a weighted sum of a set or candidate modules outputs, which is similar to our method, with two important differences: a) in the works by \cite{ma2019snr, maziarz2019gumbel}, the output is computed as a linear combination of candidates outputs, not module weights, and b) the module candidates in their methods are tied to the specific location in the network. \cite{maninis2019attentive} added a task-specific residual adapters to specialize the feature extractor to each task, The method proposed by \cite{purushwalkam2019task} performs zero-shot multitask learning by finding a task-specific routing using a gating mechanism. This approach is somewhat similar to ours, but is less efficient in terms of the number of parameters since a) the method in \cite{purushwalkam2019task} uses a modular network on top of a classical ResNet \cite{he2016deep} while our method uses only about 100 additional parameters and b) the modules used in \cite{purushwalkam2019task} are layer-specific and not reusable while there is no such constraint in our architecture.
    One important direction of study involved multitask parameter sharing through tensor factorization \cite{yang2016deep,zhao2017tensor,bulat2020incremental}. Our method can be viewed as a special case of tensor factorization with a few small yet important differences: 1) our method is based on the isometric ResNet as opposed to the regular ResNet backbone, which allows to change the order of residual blocks and scale the model by adding or removing the blocks fairly easily; 2) our method has an additional softmax-constraint on the task-specific parameters which turns out to play an important role in this architecture; 3) it enables weight sharing within the same task; and 4) we show that our compositional model can be easily applied for various applications, including transfer learning, multitask learning and domain adaptation. 
     \cite{wu2018blockdrop, newell2019feature, guo2020learning} perform task-specific model compression.
     There also exists a number of studies on the effectiveness of modular networks in the context of visual question answering. One of the early papers \cite{andreas2016neural} used a natural language parser to determine the layout of the composite network consisting of predefined modules that solve different kinds of subtasks, such as find, measure, describe etc; further works improved this approach by switching from an external parser to a fixed \cite{hu2017modeling} or arbitrary query structure \cite{hu2017learning, hu2018explainable, pahuja2019structure}.

\section{Compositional Models and Module Mixtures}
\label{sec:comp_model}

\subsection{Compositional Models}
\label{sec:conversational}

    \begin{figure*}
        \centering
        \vspace{-1.5cm}
        \includegraphics[width=1.9\columnwidth]{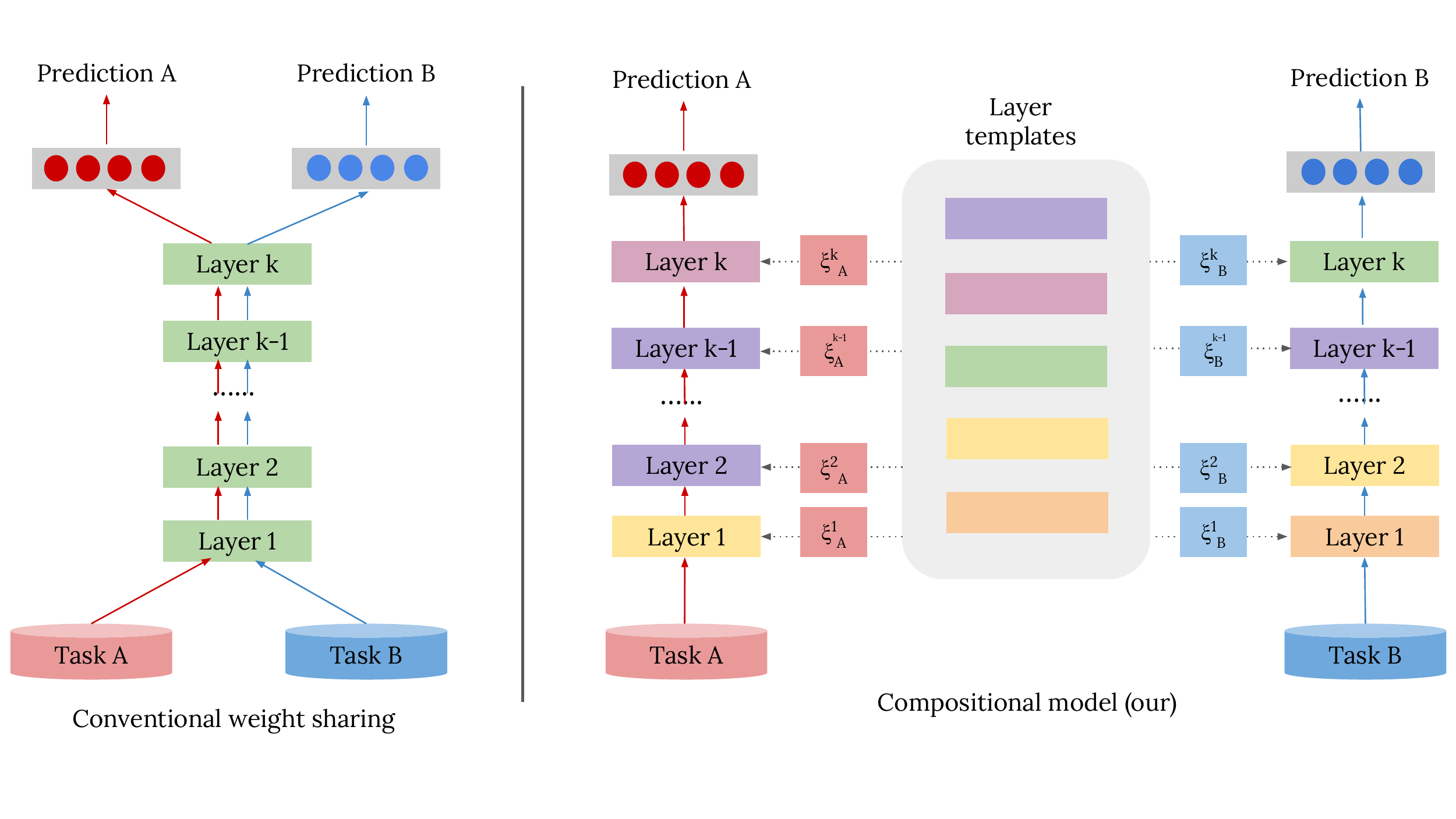}
        \caption{\textbf{Left:} conventional hard weight sharing approach. The model consists of a shared feature extractor followed by the task-specific task solvers (e.g. logits layer in case of classification). The only place where the model can specialize to each task is in its solver, which might not be enough for sufficiently distinct tasks. \textbf{Right:} Compositional model. We use a shared set of trainable modules (we call them {\em templates}) available for all tasks; each layer's weights are generated as a linear combination of the templates weights, and the linear combination weights ({\em mixture weights} $\xi$) are task-specific. In this case, the templates can not only be shared across different tasks, they also can be reused multiple times within the same task. (best viewed in color).}
        \label{fig:figure_1}
    \end{figure*}

    In conventional convolutional deep neural networks, most layers are different from each other and do not conform to a particular fixed design.
    Various model blocks may have different resolutions and different numbers of filters, some blocks may have or lack a residual connection and so on.
    This makes most model activations at different layers incompatible with each other.
    Recently, it has been demonstrated \cite{sandler:19} that high performing convolutional networks can nevertheless be composed of identical blocks.
    Such networks that were called {\em isometric networks} essentially iterate on the same activation space.
    The architecture of an isometric network includes the following core components: {\bf (a)} {\em input adapter} that maps model input $i\in I$ into the activation space $Z$; {\bf (b)} {\em model body}, a sequence of {\em blocks} all sharing the same architecture and mapping the space $Z$ to itself; {\bf (c)} {\em output adapter} mapping the output of the last block into the embedding space $E$, and finally {\bf (d)} {\em logits layer} for classification models, mapping embedding space $E$ into the final predicted probabilities.
    
    While isometric networks achieve near SOTA performance, the fact that they iterate on the same internal activation space $Z$ makes them a perfect candidate for studying {\em modular architectures}.
    Specifically, instead of treating each block as an independently trainable mapping, we may view the entire model body as a composition of modules $f_k$ each of which may appear in a sequence more than once.
    In the following, we will denote a set of trainable modules as $\F = \{f_1,f_2,\dots,f_K\}$ and their composition $f=f_{m_L}\circ \dots \circ f_{m_1}$ forming a model body\footnote{with $L$ is the total number of layers in the model} will frequently be denoted as $f_{(m_1,\dots,m_L)}$ with $(m_1,\dots,m_L)$ being called a {\em signature} of this model.
    
\subsection{Learning Compositional Models}
    
    \paragraph{Mixture modules}
    Compositional models discussed in Section~\ref{sec:conversational} can be trained on a particular task for any predefined choice of the model signature.
    But, while the study of hand-designed modular architectures is an interesting exploration in itself, one may want to go further and identify a sequence of modules $(m_1,m_2,\dots,m_L)$ together with the modules $\F$ themselves that is the most advantageous choice for solving a particular problem.
    In some previous explorations of modular architectures, authors used a probabilistic approach where the model body $(m_1,\dots,m_L)$ was sampled for every input batch or sample from the optimized probability distribution $p(m_1,m_2,\dots,m_L)$.
    For example, in \citet{kirsch:18}, modules\footnote{here each $m_k$ is actually a set of several chosen modules the output of which is combined} were sampled from a distribution defined by per-layer conditional probabilities $p(m_\ell|x_{\ell-1})$ with $x_{\ell-1}$ being the output of the previous layer ($\ell-1$), i.e., $x_{\ell-1}=f_{(m_1,\dots,m_{\ell-1})}(i)$.
    Since discrete nature of such a choice limits the flexibility of the model and since in practice learning such a model may be complicated due to high-variance estimator noise, we instead replace a composition of blocks from $\F$ with a composition of {\em mixture modules}, each of which is essentially a smooth superposition of blocks from $\F$.
    More precisely, each mixture module $h_{\bm{\xi}}$ with $\bm{\xi}\in \R^K$ has the same architecture as blocks from $\F$, but every parameter $\theta$ in this mixture module (be that some convolutional kernel, or a batch normalization $\beta$ or $\gamma$ parameter) is a linear combination $\sum_{k=1}^{K} \xi_k \theta_k$ of the corresponding parameters $\theta_k$ from $f_k\in \F$.
    Notice that since each block contains multiple convolutions with different nonlinearities, a mixture module obtained by linearly combining weights from $\F$ is not a linear combination of modules from $\F$.
    It is true however that if $\xi_i=\delta_{ij}$ for some fixed\footnote{Here $\delta_{ij}$ is a Kronecker delta} $j$ (which we later refer to as ``one-hot'' vectors), a mixture module $h_\xi$ becomes identical to $f_j\in \F$.
    Due to the fact that $h_\xi$ mixes parameters from $\F$, we will sometimes refer to blocks from $\F$ as {\em templates}, which are mixed together to finally form an actual unit of computation.
    
    After we choose a particular task and replace the model body with a composition of mixture modules $h_{\bm{\xi}_L}\circ \dots \circ h_{\bm{\xi}_1}$, we obtain a differentiable objective that can be optimized with respect to a {\em model signature} $\Xi=(\bm{\xi}_1,\dots,\bm{\xi}_L)$ and parameters of $\F$.
    In our experiments, we constrained each of $\xi_\ell$ by using a softmax nonlinearity, which allowed us to guarantee that $\sum_i (\xi_\ell)_i=1$ for every layer $\ell$.
    Of course, even with this choice of the nonlinearity, we are not guaranteed that the final signature $\Xi$ will contain only one-hot vectors $\xi_\ell$ that can be interpreted as blocks from $\F$.
    Notice that while it is possible to use other nonlinearities like $\tanh$ or not use any nonlinearities at all (analogous to a conventional simple unconstrained tensor factorization), we observed that none of such alternative choices resulted in stable high-performing models thus making softmax a crucial choice in our model.
    
    \paragraph{Regularization of mixture weights}
    While in our experiments, mixture weights $\xi_\ell$ frequently approached a set of approximately one-hot vectors\footnote{i.e., a state with $\max_i (\xi_\ell)_i \approx 1$ for most layers $\ell$}, one can use additional regularizers to ensure that all $\xi_\ell$ are one-hot and the resulting model can be represented as a composition of individual template modules.
    Here we outline two choices that we experimented with: (a) entropy regularization that penalizes $\max_i (\xi_\ell)_i < 1$, but does not involve template weights and (b) a regularizer that penalizes $\Xi$ and also effectively attracts individual templates towards the generated mixture weights to which they contribute the most.
    
    \paragraph{Entropy regularization}
    Treating each $\xi_\ell$ as a probability distribution, we can minimize entropy $H[\xi_\ell]$ to regularize the mixture weights vectors $\xi_\ell$:
    \begin{equation*}
        H[\xi] = - \sum_{k=1}^{K} \xi_k \log \xi_k.
    \end{equation*}
    However, this regularizer has one significant flaw: if $\xi_i < \xi_j$ for some $i$ and $j$, the gradient of $H[\xi]$ will push $\xi_i$ and $\xi_j$ to move further apart thus preventing them from switching their order and reaching a state where $\xi_i > \xi_j$.
    
    \paragraph{``Clustering'' regularizer}
    To avoid this issue, in all our experiments, we used another regularizer $R[\xi,\theta]$:
    \begin{equation*}
        R[\xi,\theta] = \alpha \sum_{k=1}^{K} \xi_k g(\|\bar{\theta} - \theta_k\|),
    \end{equation*}
    where $\bar{\theta} = \sum \xi_k \theta_k$, $g$ is an arbitrary monotonically increasing function with $g(0)=0$, $\alpha$ is a regularization weight and $\theta_k$ are parameters of a block $f_k \in \F$.
    It is easy to see that when all $\theta_k$ are different, $R[\xi,\theta]$ is minimized for one-hot vectors $\xi$.
    At the same time, the gradients of $R$ with respect to $\theta_k$ push templates towards the current average weight $\bar{\theta}$ with an effective force proportional to $\xi_k$.
    Intuitively, these potentially conflicting ``clustering'' forces getting stronger for templates contributing the most to a particular layer parameters, together with the force making mixture weights more singular, should allow the templates to self-organize.

    \paragraph{Weight generation interpretation}
    \def\FF{\mathfrak{F}}
    Consider a single block of our isometric network and let $\FF$ be a function family containing single-block tensor transformations corresponding to all possible choices of block weights.
    All neural networks obtained by composing $L$ such blocks without any weight-tying form a function family $\FF^L$.
    Now, instead, consider a model composed of mixture-weight blocks with templates from $\F$.
    In this case, the family of all possible single-block transformations is only controlled by a vector of mixture weights $\bm{\xi}$ and narrows down to a family further denoted as $\FF_{\F}$.
    A composition of mixture-weights blocks is thus $\FF_{\F}^L$.
    In a sense, choosing templates $\F$, we introduce a small function family that can be parameterized by just a handful of real-valued parameters $\Xi$.
    This perspective shows a connection of the proposed approach to other approaches parameterizing neural network weights like, for example, \textsc{HyperNetworks} \citep{ha2017hyper}.
    
\section{Applications and Experimental Results}
    \label{sec:Experiments}

\subsection{Isometric Model Details}
    \label{sec:Model}

    Before discussing our experimental results, we first need to briefly describe the architecture of the Isometric network used as a basis of these experiments.
    As described in Section~\ref{sec:conversational}, our implementation of the modular Isometric network contains the following four components (see Appendix~\ref{app:model} for details): {\bf (a)} {\em input adapter}, a nonlinear convolution with the kernel and the stride of the same size (typically 4 to 16), which effectively downsamples the image while increasing the number of channels in the tensor to $40\varrho$, where $\varrho$ is the {\em model depth multiplier} controlling the complexity of the network; {\bf (b)} {\em model block}, a typical MobileNetV3 \cite{howard2019mobilenet}, or Isometric model residual block \cite{sandler:19} with the ``expansion'', depthwise and ``projection'' convolutions and a squeeze-and-excite block \cite{hu:2020}; {\bf (c)} {\em output adapter}, a sequence of two nonlinear kernel-1 convolutions with an average pooling layer in between and finally {\bf (d)} {\em logits layer}, a fully-connected layer producing model predictions.
    In the following, we will rely on two specific designs: (a) larger design more suitable for complex tasks that uses $40$-channel tensors (or $\varrho=1$) and the embedding size of $1280$ (as produced by the output adapter) and (b) smaller design typically using $24$-channel tensors\footnote{We used the number of channels divisible by $8$ for training efficiency.} and the embedding size of $256$.

    In our implementation of networks with mixture modules, we used mixture weights $\xi\in \R^K$ for mixing all variables appearing in every component of the block except for batch normalization (BN) parameters: this included all convolutional weights and biases, and all components of the squeeze-and-excite block.
    In some experiments, we also used the same $\xi$ to produce a linear combination of batch normalization $\beta$ and $\gamma$ parameters (but never moving mean variables).
    In the following, we will refer to such experiments as experiments with ``mixed batch normalization parameters''.
    
    Notice that generating block parameters from templates and mixture weights can be expensive for a large number of templates in $\F$.
    However, since in this work we perform this calculation for an entire batch and not for individual samples, this procedure did not introduce a significant computational overhead.

\subsection{Learning a Single Task}
\label{sec:single-task}

    We began our study of networks with mixture weights by training these models on individual supervised tasks.
    In our initial experiments, we studied models with $4$ to $32$ layers and $2$ to $16$ templates forming $\F$.
    We trained these models on the \textsc{CIFAR-100} dataset using a 24-channel tensors and a smaller output adapter (see Appendix~\ref{app:model}).
    
    When training a model on a single task, we observe that even in the absence of regularization, the mixture weights $\Xi$ frequently approach a collection of one-hot vectors, i.e., $\max_i (\xi_\ell)_i \approx 1$ for most layers $\ell$.
    Discovered model signatures (with one regularized example having a signature $(1,2,2,3,3,3,3,4,4,4,4,4)$) are discussed in Appendix~\ref{app:single-task} in detail.
    After studying multiple such signatures, we observed that they tend to repeat the same modules in contiguous uninterrupted sequences and the number of repetitions grows towards the end of the model.
    In other words, early modules use fewer repetitions.
    We call such pattern {\em progressive}.

    Once the optimal signature is identified, we investigated  whether this particular model architecture is truly optimal for solving a particular task.
    We compared networks with mixture weights to those with hand-designed signatures of three types: {\bf (a)} {\em sequential signatures} $(f_K^{L/K}) \circ \cdots \circ (f_1^{L/K})$ repeating each of $K$ templates $L/K$ times, where $L$ is the number of layers; {\bf (b)} {\em cyclic signatures} corresponding to $(f_K\circ \cdots \circ f_1)^{L/K}$ and {\bf (c)} random signatures.
    Our experimental results (for details see Appendix~\ref{app:single-task}) appear to indicate that mixture-weight results are comparable to those obtained with a sequential signature and overall outperform results for other considered model signatures.
    However, it remains an open question whether progressive signatures are actually optimal, or are an artifact of the modular network training.
    
    In Table~\ref{tab:imagenet-results} we show our results of improving \textsc{ImageNet} accuracy by increasing the number of layers while keeping the number of parameters intact.
    Due to hardware and time limitations we only experiment with progressive pattern and leave more detailed study for future work.
    More details on \textsc{ImageNet} architecture is included in the Appendix.
    
    \begin{table}
        \centering
        \caption{
            \small \textsc{ImageNet} results, comparing performance of different module sequences.
            We use 8 and 16-layer isometric models from \cite{sandler:19}.
            For our method we use the same models, but the modules are arranged into $32$ and $48$ layers as described in Section~\ref{sec:single-task}.
            The models are of the same size as baseline.
        }
        \inctabcolsep{-2pt} { 
        \begin{tabular}{cccc}
            {\small\bf Isometric-8} & {\small\bf 8/16 model}              & \bf{\small Isometric-16} & \bf{\small 16/48 model} \\
            \hline
                68.3            &  \bf{70.6} (+2.3\%)     &  71.7    &  \bf{72.8} (+1.1\%)   \\
            \hline
        \end{tabular}
        }
        \label{tab:imagenet-results}
    \end{table}

\subsection{Multi-Task and Transfer Learning}

    \begin{figure*}
        \centering
        \includegraphics[width=0.9\linewidth]{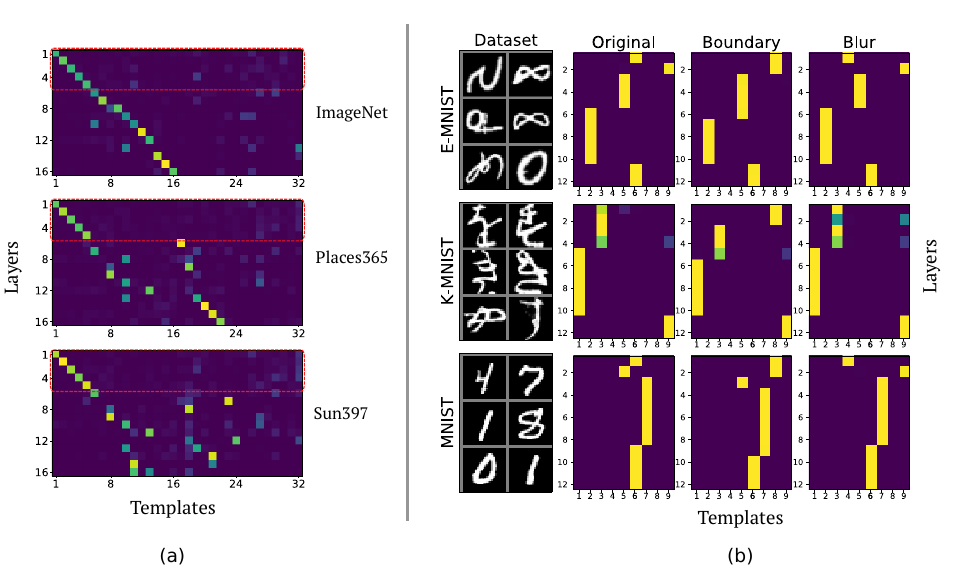} 
        \caption{
            Self-organization of modules for multi-task and multi-domain learning.
            In (a) co-training $5$ natural image datasets with $32$ modules (\textsc{CIFAR-100} and \textsc{Food101} not shown), first 5 layers were reused across all learned models (shown with dashed rectangles).
            In (b) co-training $9$ modules on $9$ tasks: $3$ handwritten digit datasets \textsc{EMNIST}, \textsc{MNIST} and \textsc{KMNIST} with the original images and two image augmentations (boundary extraction and blurring).
            The model learned to use different input transformation for each image augmentation, while sharing the top layers for each task. On the other hand, for each fixed augmentation, the model learned to share first few layers across all different tasks. 
        }
        \label{fig:organization}
    \end{figure*}

    \paragraph{Multi-task learning}
    Another potential application of compositional models is in multi-task learning, where we expect the computation to be partially shared across different tasks.
    In this application, each task $T_\nu$ uses its own private set of mixture weights $\Xi_\nu$, but the template blocks $\F$ and input, output adapters are shared across all tasks.
    As the number of tasks grows, the total number of blocks in $\F$ should also be scaled up to achieve sufficiently high accuracy on all $\{T_\nu\}$.
    % \textcolor{gray}{However, we hypothesize that the average number of blocks per task goes down in the process should templates be meaningfully reused across tasks.} \textcolor{blue}{Is it backed up by experimental results? If not, better not mention.}
    Also note that modular networks provide a convenient tool for training sets of tasks parameterized by additional arguments like those, for example, used in basic visual question answering modules (finding specific objects, transforming and filtering attention regions, etc.) \citep{hu2018explainable,hu2017learning}.

    Notice that in its current form, this approach is a generalization of \textsc{AdaShare} \citep{sun2019adashare}, in which reused modules always retain their relative order with respect to each other\footnote{and cannot be reused later in the same network}.
    Since multi-task training is not the primary application explored in this paper and the direct comparison is not possible without re-implementing the method, we do not compare our results with \textsc{AdaShare}.

    While the tasks $\{T_\nu\}$ could be very diverse and along with conventional supervised tasks could also include unsupervised tasks, or even generative and image-to-image translation objectives, here we will only consider a collection of image classification tasks.
    For our experiments, we chose a collection of natural image datasets including: \textsc{ImageNet}, \textsc{CIFAR-100}, \textsc{Places365}, \textsc{Sun397} and \textsc{Food101}.
    Here we used model depth multiplier $\varrho$ of $1$ and a larger output adapter (see Section~\ref{sec:Model}).
    We compared multiple different ways of co-training different classification tasks: {\bf (a)} baseline model with or without per-task batch normalization $\beta$ and $\gamma$ variables (see {\em linear patches} in \citet{mudrakarta:19}) and {\bf (b)} model with mixture-weights with or without per-task batch normalization variables.

    In our experiments summarized in Appendix~\ref{app:multitask} (Table~\ref{tab:transfer_ext}), we noticed that the mixture-weight model with $K=32$ templates achieved a lower overall cross-entropy loss and thus fitted training data better than the baseline models.
    This did translate to a superior \textsc{ImageNet} performance, but in some cases resulted in worse accuracies for simpler datasets, on which overfitting seemed to occur.
    
    \paragraph{Self-Organization of Computational Blocks}
    Perhaps one the most interesting aspects of networks trained with mixture modules is that they essentially involve self-organization of atomic computational blocks.
    The properties of emergent block configurations are insightful both when we train a model on a single task and when performing multi-task learning.

    In the case of a multi-task learning, we discover that the locations of the blocks re-used across tasks agrees with our intuition regarding the nature of the task similarity.
    Specifically, if all tasks operate on a similar domain, first few layers of all models appear to share the same mixture weights (see Figure~\ref{fig:organization}(a)) thus sharing early-stage image processing.
    For similar tasks operating on different image domains, we see the opposite situation (see Figure~\ref{fig:organization}(b)), where the first few layers differ, but the late sample processing is shared across all tasks performing inference on the same dataset (\textsc{EMNIST}, \textsc{MNIST} and \textsc{KMNIST} in our experiments).
    In this case, first few layers essentially adapt network input to a particular common format shared across all related tasks and processed by the following layers.

    A set of blocks $\F$ pretrained on a collection of tasks can later be used for transferring extracted knowledge onto a new task.
    Adjusting mixture weights to form a new model potentially gives us more flexibility than tuning logits or batch normalization parameters alone.

    \paragraph{Transfer learning}
     \begin{table}
        \centering
        \caption{
            Transfer learning results from model trained on multiple initial datasets (\textsc{ImageNet}, \textsc{CIFAR-100}, \textsc{Places365}, \textsc{Sun397} and \textsc{Food101}) onto new datasets.
            For baseline column we train multi-task model following \cite{mudrakarta:19}, with shared backbone and per-task logits and batch normalization parameters.
            We then transfer to the target datasets by fine-tuning logits and batch normalization parameters.
            Other columns show results obtained by our method, where we use equivalent architecture, but with mixture weights both for initial model and each transfer task.
            Each column shows results with mixture coefficients initialized from the corresponding training task.
            Our method adds less than $500$ additional parameters for each transfer task.
        }
    \inctabcolsep{-2pt} { 
        \begin{tabular}{ccccc}
        \textbf{\small Dataset} & {\small\bf Baseline} & \bfseries{\small\scshape{ImageNet}} & \bfseries{\small\scshape{CIFAR}} & \bfseries{\small\scshape{Food}} \\
        \hline
        \textsc{Aircraft} & $43.3$         & $\mathbf{43.9}$ & $42.6$ & $39.4$ \\
        \textsc{Cars} & $49.2$             & $\mathbf{51.0}$ & $46.7$ & $42.2$ \\
        \textsc{EMNIST} & $\mathbf{85.5}$  & $85.0$ & $85.2$ & $83.6$ \\
        \textsc{Stanford Dogs} & $55.6$    & $\mathbf{57.6}$ & $42.4$ & $36.5$ \\
        \hline
        \end{tabular}
        }
        \label{tab:transfer}
    \end{table}
    
    In our experiments, we used models pretrained on 5 datasets: \textsc{ImageNet}, \textsc{CIFAR-100}, \textsc{Places365}, \textsc{Sun397} and \textsc{Food101} (see Appendix~\ref{app:multitask}).
    We then used a learned collection of templates $\F$ for transferring knowledge to different new tasks such as \textsc{Stanford Dogs}, \textsc{Aircraft}, \textsc{Cars196} and \textsc{EMNIST}.
    % In our experiments, we either used just the original templates $\F$, or introduced additional trainable templates.
    We compared our results with the baseline obtained by fine-tuning logits and BN parameters of a baseline model trained with a common model backbone, but per-task logit layers and BN parameters (``Baseline with patches'' in Table~\ref{tab:multi}).
    Proper initialization of mixture weights $\Xi$ proved to be extremely important for achieving best performance.
    Model initialized with random $\Xi$ performed significantly worse than the models initialized with $\Xi$ corresponding to one of the datasets in the original mixture.
    In the following, we will denote by $\Xi_\nu$ a set of mixture weights used for a task $\nu$ in the original mixture-module network trained on a collection of datasets.
    Initializing the model at some particular $\Xi_\nu$, we compared model performance with and without fine-tuning $\Xi$ (fine-tuning BN parameters and logits in both cases).
    
    The final results are presented in Table~\ref{tab:transfer}.
    Experimental observations suggest that a proper choice of the dataset $\nu$ for $\Xi=\Xi_\nu$ is extremely important: using $\Xi$ values from the \textsc{Food196} model resulted in a model that achieved the same $>99\%$ training accuracy, but would frequently lead to a $10\%-20\%$ smaller validation accuracy.
    We also observe that the validation accuracy improvement in the final model can be partly attributed to a higher validation accuracy on the chosen model $\Xi_\nu$ and partly to the result of fine-tuning this $\Xi\approx \Xi_\nu$ value.

\subsection{Domain Adaptation}
\label{sec:uda}
    
    In this section, we explore the use of compositional models for unsupervised multi-source domain adaptation which aims to minimize the domain shift between multiple labeled source domains and an unlabeled target domain. To our knowledge, we are the first to apply modular networks for domain adaptation and to show the efficiency and limitations of the modular paradigm for domain transfer.
    To illustrate the ability of compositional networks to perform distribution alignment, we adopted two existing unsupervised domain adaptation methods: Adversarial Discriminative Domain Adaptation (ADDA) by \cite{tzeng2017adversarial} and Moment Matching for Multi-Source Domain Adaptation (M3SDA) by \cite{peng2019moment}.
    We performed experiments with three commonly used domain adaptation datasets: digits datasets (MNIST, corrupted MNIST, SVHN and USPS), Office-Caltech10 and DomainNet\cite{peng2019moment}.
    The code used for the domain adaptation experiments can be found on our project page (\textit{URL will be here}). % (\url{https://github.com/googleinterns/loop-project}).
    The details of implementation of the compositional domain adaptation and a review of related work on domain adaptation can be found in the Appendix~\ref{apx:uda}.
    
    \paragraph{Multi-source adversarial classification} In the compositional adversarial classification method, the shared classifier is trained to perform accurate classification on all source datasets. Additionally, a domain discriminator is trained to distinguish between the late features of source and target domains (in our implementation, the output of the layer before the output adapter). As in the multitask learning scenario, the model contains separate mixture weights for each domain; the mixture weights of source domains are trained along with templates to minimize the classification loss, while the target domain mixture weights are trained to "fool" the domain discriminator by aligning the target feature with the source features. 
  
    \begin{table}[ht]
        \centering
        \caption{Multi-source unsupervised domain adaptation target test accuracy results with compositional models on Digits domains (MNIST, corrupted MNIST (shear), corrupted MNIST (scale), corrupted MNIST (shot noise), SVHN $\rightarrow$ USPS), Office-Caltech10 (amazon, caltech, dslr $\rightarrow$ webcam) and a subset of 100 most frequently occurring classes of DomainNet domains (clipart, infograph, quickdraw, painting, real $\rightarrow$ sketch). Baseline methods: Adversarial Discriminative method and Multi-Source Moment Matching method. We pretrained the baseline isometric model on source domains and fine-tuned only the input adapter (IA) containing $\approx 1000$ parameters. We applied the same domain adaptation techniques for our compositional model (``Comp.'') and fine-tuned only mixture weights (MW) consisting of $\approx 100$ parameters. We included the result of supervised mixture weights finetuning to upper bound the target accuracy achievable by unsupervised adaptation of mixture weights.
        }
        \vspace{3mm}
        \inctabcolsep{-2pt} { 

        \begin{tabular}{lcccc}
        {\bf Method} & {\bf Digits}  & {\small\bf Office-C10} & {\small\bf DomainNet} \\ 
        \hline
        Source-only & $95.75$  & $42.96$ & $24.75$ \\
        Comp. source-only & $96.05$ & $41.89$ & $11.61$ \\
        Adversarial (IA) & $98.12$ & $\bm{57.79}$ & $\bm{27.54}$ \\
        Comp. Adversarial  & $\bm{98.33}$ & $48.18$ & $14.40$ \\
        Moment Matching (IA) & $96.93$ & $51.39$ & $26.12$ \\
        Comp. Moment Matching  & $96.85$ & $52.76$ & $14.22$ \\
        \hline     
        Comp. supervised & $98.25$ & $53.57$ & $15.46$ \\
        \hline   
        \end{tabular}
        }
        \label{tab:adaptation}
    \end{table}
    
    \begin{table}[ht]
        \centering
        \caption{
        Supervised training test accuracy of a standard isometric model and compositional model with separate mixture weights for DomainNet domains. The models were pretrained on the clipart, infograph, painting, quickdraw and real domains; results include source-only accuracy on sketch domain. 
        }
        \begin{tabular}{l c c c c}
        {\bf Domain} & {\bf Isometric}  &  {\bf Compositional} \\ 
        \hline
        clipart   & $75.98$   & $\bm{79.37}$\\
        infograph & $23.62$   & $\bm{50.38}$\\
        painting  & $44.17$   & $\bm{52.80}$\\
        quickdraw & $28.97$   & $\bm{58.54}$\\
        real      & $\bm{54.04}$   &$53.66$ \\
        \hline
        sketch source-only & $\bm{26.17}$ & $11.61$\\
        sketch finetuning  & $\bm{27.54}$ & $15.46$\\
        \hline
        \end{tabular}
        \label{tab:sup_adaptation}
    \end{table}
    
    \paragraph{Moment Matching}
    As in the original paper introduced by \cite{peng2019moment}, the method performs domain adaptation by minimizing the distance between the source and target domain statistics (more details can be found in the original paper). In the compositional moment matching setting, only target domain mixture weights are trained to perform moment matching.  In addition to that, the model has two logits layer instead of one which are used to perform minmax discrepancy optimization (see \cite{peng2019moment} for more detail). In the compositional case, only the mixture weights are affected by the discrepancy minimization. 
    
    \paragraph{Results}
    As we can see in the Table \ref{tab:adaptation}, fine-tuning only the mixture weights for the domain adaptation objective improves the classification accuracy on target domain. To compare the results of compositional domain adaptation, we performed adversarial and moment matching adaptation with an isometric model without weight sharing as a backbone, and for fair comparison we only fine-tuned the input adapter (\cite{saito2019strong} point out that finetuning the first layers is appropriate for the case when the domain shift is mostly in the low-level visual features). Results in the Table \ref{tab:adaptation} on all three datasets indicate that the upper bound on the target accuracy that can be achieved by finetuning the mixture weights is significantly lower than the one that can be achieved by finetuning the non-shared isometric model. This can be explained by the fact that the architecture of the non-modular networks, such as the isometric model, enforces learning of the features common across all domains, whereas the compositional architecture allows specification of the layers weights to each specific domain. This intuition can be further supported with the results in Table \ref{tab:sup_adaptation} that show test accuracy of the isometric model and the compositional model on the source domains they were trained on and the source-only accuracy on target domain. The fact that accuracy on all source domains are significantly larger for the compositional model and the opposite is observed on the target accuracy, suggests that compositional models are able to specify the parameters to each source domain, which means that the model does not need to learn features common across domains which results in worse performance on the unseen domain. Similar results for Digits and Office-Caltech10 domains can be found in section \ref{sec:sup_multi_da} of the Appendix. 

\section{Discussion}
\label{sec:discussion}
    In this paper, we introduced a novel approach to modular neural networks that decouples modules from their position in the network thus allowing reuse of modules not only across tasks, but also across layers.
    Our method beats the baseline on multitask learning with additional templates and achieves better transfer learning results on most of the datasets we explored, which we partly attribute to less aggressive parameter over-sharing and partly to the ability of our model to fine-tune the module order itself.
    We then apply our method to training parameter-efficient networks and show that we can scale SOTA \textsc{ImageNet} classification models to achieve higher accuracy while keeping the number of parameters fixed.
    We also show that simple fine-tuning of mixture weights allows unsupervised domain adaptation improving model's performance on target domain by a large margin.
    \\
    Since, by design, the compositional models divides the task into simple subtasks, it can provide greater understanding of the computation process.
    For example, we notice that co-training multiple tasks on the same image domain, all trained models share early image processing modules.
    At the same time, when the dataset contains multiple image domains of the otherwise identical task, early layers bring network activations to the common format, while the final stage of the computation defined by the sequence of participating modules is shared.

{\small
    \bibliographystyle{plainnat}
    \bibliography{ms}
}

\newpage

\appendix

\section*{Supplementary Materials}

    \section{Model and Training Parameters}
\label{app:params}

\subsection{Model Details}
\label{app:model}

    Here we provide additional details about the models that were used in our experiments.
    The approximate number of parameters in individual model components are shown in Table~\ref{tab:params}.

    \paragraph{Model blocks}
        Just like in the original isometric model \citep{sandler:19}, each residual block used the expansion factor of $6$, i.e., ``expansion'' convolution mapped input tensor with $40\varrho$ channels into the intermediate $240\varrho$-channel tensor, which was later projected back into the tensor with $40\varrho$ channels.
        The intermediate depthwise convolution used the kernel size of $3$.
    \paragraph{Output adapter}
        Output adapter was chosen to be a sequence of two nonlinear kernel-1 convolutions with an average pooling layer in between.
        Assuming that the first convolution produces a tensor with $c_1$ channels and the second convolution produces a tensor with $c_2$ channels, the total number of parameters in this part of the model was approximately $40\varrho c_1 + c_1 c_2 + c_1 + c_2$.
        For larger networks, we used $c_1=960$ and $c_2=1280$, while for smaller networks we chose $c_1=128$ and $c_2=256$.
    \paragraph{Logits layer}
        Final fully-connected layer mapped embedding vectors produced by the output adapter into the final predictions.
        It used about $c_2 C + C$ parameters with $C$ being the total number of labels.
    
\subsection{Training Parameters}
\label{app:TrainParams}

    In all our experiments, we used \textsc{RMSProp} optimizer.
    In smaller scale experiments, the learning rate was chosen to be either $0.02$, or $0.04$.
    The dropout keep probability and the weight decay were $80\%$ and $10^{-5}$ correspondingly.
    In all of our experiments, we used exponential moving averages of all trained variable for inference.
    
    All datasets that we experimented with are available in {\tt tensorflow datasets}.
    For the {\tt airplane} dataset, we used an $80\%$ of the total number of labeled samples for the training set and $20\%$ of samples for the test set.

\section{Single-Task Learning}
\label{app:single-task}

    In Section~\ref{sec:single-task}, we explored single-task training of networks with mixture modules.
    Model signatures that we discovered in the process contained multiple repeated occurrences of the same template with the number of repetitions increasing towards the tail of the model.
    But are these discovered signatures optimal and if so, how much better are they compared to other possible signatures?
    Here we discuss experiments conducted for several hand-designed model signatures and study how model performance changes if we perturb the sequence of modules and perform inference using a different order of computational blocks compared to the one the models have been trained with.

\subsection{Models with Different Signatures}

\begin{figure}
        \centering
        \includegraphics[width=0.45\textwidth]{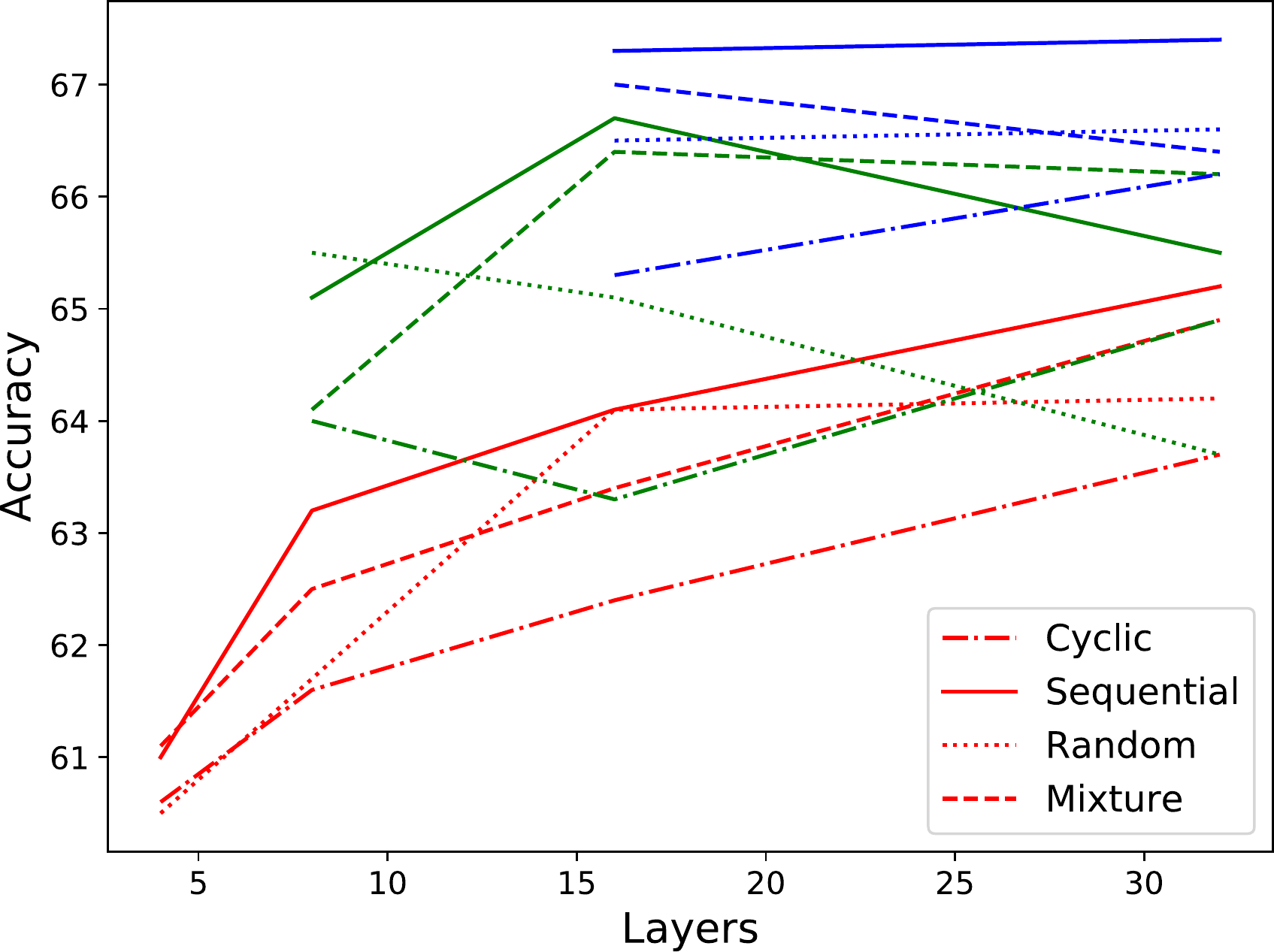} 
        \caption{
            Visualization of results with fixed and trainable signatures on \textsc{CIFAR-100} presented in Table~\ref{tab:SingleTask}: validation accuracies for signatures with 2 templates (red), 4 templates (green) and 8 templates (blue).
        }
        \label{fig:SingleTask}
    \end{figure}

    In all experiments discussed here, we train and evaluate classification models on the \textsc{CIFAR-100} dataset.
    We explore models with different signatures: {\bf (a)} {\em sequential signatures} $(f_K^{L/K}) \circ \cdots \circ (f_1^{L/K})$ repeating each of $K$ templates $L/K$ times, where $L$ is the number of layers; {\bf (b)} {\em cyclic signatures} corresponding to $(f_K\circ \cdots \circ f_1)^{L/K}$, {\bf (c)} {\em random signatures}, and {\bf (d)} networks with mixture modules.
    In all experiments, we chose $\varrho=0.6$ and a smaller output adapter (see Appendix~\ref{app:model}).
    Specific model and training parameters chosen for performing these experiments are outlined in Appendix~\ref{app:params}.
    Experimental results can be found in Table~\ref{tab:SingleTask} and Figure~\ref{fig:SingleTask} shows validation accuracies for different signature choices, different numbers of modules $K$ in $\F$ and different numbers of layers $L$.
    While we did not have an opportunity to gather sufficient statistics for all of these experiments, reproducing individual experiments we observed accuracy fluctuations with a standard deviation of approximately $0.5$.
    
\begin{figure}[t]
        \centering
        \includegraphics[width=0.49\textwidth]{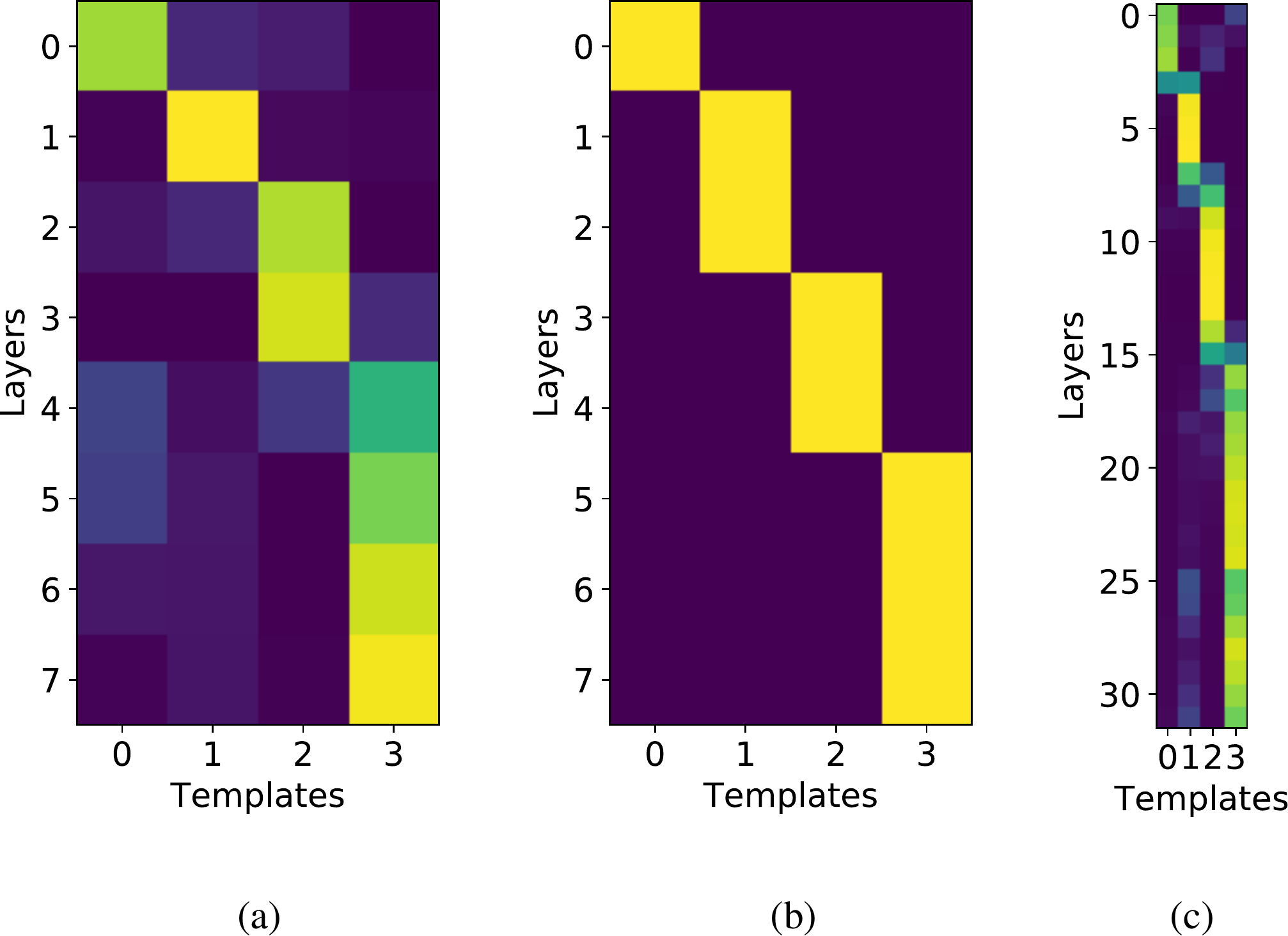} 
        \caption{
            Examples of mixture-weight model signatures obtained with 4 modules and 8 layers: (a) mixture weights mid-training without any extra regularizers; (b) mixture weights obtained with the ``clustering'' regularizer; (c) mixture weight for a 32-layer model with 4 templates.
        }
        \label{fig:signatures}
    \end{figure}

    As one would expect, we see that the model accuracy grows monotonically both with $K$ and $L$.
    Sequential signatures are seen to outperform models with cyclic and random signatures.
    At the same time, models with mixture weights appear to have an accuracy close to that of sequential models.
    Examples of discovered mixture weights are shown in Figure~\ref{fig:signatures}.
    After studying multiple such signatures, we observed that they tend to repeat the same modules in contiguous uninterrupted sequences and the number of repetitions grows towards the end of the model.
    In other words, early modules use fewer repetitions.
    Interestingly, even in the absence of regularization, the mixture weights $\Xi$ frequently approach a collection of one-hot vectors, i.e., $\max_i (\xi_\ell)_i \approx 1$ for most layers $\ell$.

    We also conducted experiments with modules containing blocks that accept both the output of the previous layer and the original input fed into the first block (output of the input adapter).
    Here the difference between different signatures shown in Table~\ref{tab:InputBlock} is even more dramatic.
    
    \begin{table}
        \centering
        \caption{
           Comparison of validation accuracies for networks with fixed and trainable signatures on the \textsc{CIFAR-100} dataset.
           We compare different types of signatures with a particular number of layers $L$ and number of templates $K$: (a) cyclic signatures [$(1, 2, 1, 2)$], (b) sequential signatures [$(1, 1, 2, 2)$], (c) random signatures (obtained by randomly shuffling a sequential or repeated signature) and (d) mixture-weight models.
           Top results (and those smaller by at most $0.5$) are highlighted.
        }
        \begin{tabular}{cccccc}
        \multirow{2}{*}{$K$} & \multirow{2}{*}{$L$} & \multicolumn{4}{c}{{\bf Signature}} \\
        & & \multicolumn{1}{c}{Cyclic} & Sequential & Random & Mixture \\
        \hline
        \multirow{4}{*}{2} & 4 & {\bf 60.6} & {\bf 61.0} & 60.5 & {\bf 61.1} \\
          & 8 & 61.6 & {\bf 63.2} & 61.7 & 62.5 \\
          & 16 & 62.4 & {\bf 64.1} & {\bf 64.1} & 63.4 \\
          & 32 & 63.7 & {\bf 65.2} & 64.2 & {\bf 64.9} \\
        \hline
        \multirow{3}{*}{4} & 8 & 64.0 & {\bf 65.1} & {\bf 65.5} & 64.1 \\
          & 16 & 63.3 & {\bf 66.7} & 65.1 & {\bf 66.4} \\
          & 32 & 64.9 & 65.5 & 63.7 & {\bf 66.2} \\
        \hline
        \multirow{2}{*}{8} & 16 & 65.3 & {\bf 67.3} & 66.5 & {\bf 67.0} \\
          & 32 & 66.2 & {\bf 67.4} & 66.6 & 66.4 \\
        \hline
        \end{tabular}
        \label{tab:SingleTask}
    \end{table}

    \begin{table}
        \centering
        \caption{
           Comparison of validation accuracies for the \textsc{CIFAR-100} dataset in models where each block receives the output of the previous layer and the original input (output of the input adapter).
           We compare different types of signatures with a particular number of layers $L$ ($12$ or $16$) and $4$ templates.
           We consider: (a) cyclic signatures [$(1, 2, 1, 2)$], (b) sequential signatures [$(1, 1, 2, 2)$], and (c) mixture-weight models.
        }
        \begin{tabular}{cccccc}
        \multirow{2}{*}{$K$} & \multirow{2}{*}{$L$} & \multicolumn{3}{c}{{\bf Signature}} \\
        & & \multicolumn{1}{c}{Cyclic} & Sequential & Mixture \\
        \hline
        \multirow{2}{*}{$4$} & $12$ & $65.7$ & $\bf{67.0}$ & $\bf{67.3}$ \\
          & $16$ & $64.9$ & $\bf{67.6}$ & $\bf{67.6}$ \\
        \hline
        \end{tabular}
        \label{tab:InputBlock}
    \end{table}

    \begin{table*}[b]
        \centering
        \caption{
            Approximate number of parameters used in each component of the model: (a) $2000\varrho$ parameters in {\em input adapter}, (b) $40\varrho c_1 + c_1 c_2 + c_1 + c_2$ in the {\em output adapter} and (c) $c_2 C + C$ in the {\em logits layer}.
            Here $\varrho$ is the model depth multiplier (affecting the number of channels), $C$ is the number of classes and $(c_1,c_2)$ are the numbers of output channels of two convolutions used in the output adapter.
        }
        \begin{tabular}{cccccc}
        & {\bf Input Adapter} & {\bf Block with SE} & {\bf Block without SE} & {\bf Output Adapter} & {\bf Logits} \\
        \hline
        Smaller ($\varrho=0.6$) & $1200$ & $20000$ & $9000$ & $37000$ & $256 C$ \\
        Larger ($\varrho=1$) & $2000$ & $52000$ & $22000$ & $1.27\cdot 10^6$ & $1280 C$ \\
        \hline
        \end{tabular}
        \label{tab:params}
    \end{table*}
    
\section{Multi-Task Learning}
\label{app:multitask}

    Here, we summarize multi-task training results obtained for a set 5 supervised tasks trained on: \textsc{ImageNet}, \textsc{CIFAR-100}, \textsc{Places365}, \textsc{Sun397} and \textsc{Food101} datasets.
    In our experiments, we compared the following multi-task training approaches:
    \begin{enumerate}
        \item {\bf Simple baseline}:
            A more straightforward approach, in which all model blocks are conventional model layers that do not use any parameter sharing (across layers).
            Here input adapter, model body and the output adapter were shared by all models solving different tasks and only the logits layers were trained independently for every head.
            Since model activations may differ for different tasks, we also accumulated per-task batch normalization moving means (adding thousands of parameters for each task on top of parameters for the final logits).
        \item {\bf Baseline with linear patches}:
            A similar approach inspired by \citet{mudrakarta:19}, where batch normalization $\beta$ and $\gamma$ variables were not shared and trained for each task independently.
        \item {\bf Model with mixture weights}:
            Model with mixture weights, where all mixture weights $\xi_i$ were initialized randomly and all batch normalization $\beta$ and $\gamma$ were also linearly mixed using these mixture weights\footnote{moving means and variances were not shared}.
        \item {\bf Model with mixture weights and linear patches}:
            Same as the previous model, except $\beta$ and $\gamma$ variables were not shared here, but were trained for each layer and each task independently.
    \end{enumerate}
    The final comparison results are presented in Table~\ref{tab:multi}.

    While it appears that additional templates result in overall performance improvement, there are reasons to believe that the model does not reach a true optimum.
    Notice that the model with linear patches can be viewed as a special case of a network with mixture modules and linear patches, where the total number of templates $K$ is the same as the number of layers $L$ and all mixture weights for all tasks satisfy $(\xi^T_\ell)_k=\delta_{\ell,k}$, where $\xi^T_\ell$ are the mixture weights for layer $\ell$ for task $T$.
    In our experiments where we chose $K=L=16$ (see Table~\ref{tab:multi}), the final training loss turned out to be higher than that of the conventional model with linear patches, which suggests that the training procedure fails to identify a solution in which every layer uses a unique template.
    As a sanity check, we verified that by initializing mixture weights $\xi_\ell$ at $(\xi_\ell)_k \approx \delta_{\ell,k}$, the resulting model matches baseline model performance (with linear patches) much better and the final cross-entropy losses are essentially identical.
    Another indication that our model does not currently reach the optimum is the fact that the models with mixture weights and linear patches appear to perform worse than the model with mixture weights alone.
    Improving modular network training will be one of priorities of our future work.

    \begin{table*}[!ht]
        \centering
        \caption{
            Comparison of models co-trained on 5 datasets.
            Models marked with ``diag'' used mixture weight initialization with $(\xi_\ell)_i \approx \delta_{i,\ell}$.
            Reported values correspond to validation accuracies with the training accuracy also specified for \textsc{ImageNet} and \textsc{Places} datasets.
            Training accuracy for other datasets reached $99.9\%$ in all experiments.
            Because of this, the final comparison of model performance may need to rely on \textsc{ImageNet} accuracy and the final aggregated cross-entropy loss $L_{\rm CE}$.
            Here $K=|\F|$ is the number of templates in $\F$ and models with ``linear patches''  \citep{mudrakarta:19} are models that maintain distinct per-task batch normalization $\beta$ and $\gamma$ variables.
        }
        \vspace{3mm}
        
        \begin{tabular}{lcccccc}
        \multicolumn{1}{c}{\multirow{2}{*}{\bf Model}} & \multicolumn{5}{c}{{\bf Dataset}} & \multirow{2}{*}{$L_{CE}$} \\ 
        & \textsc{CIFAR} & \textsc{Food} & \textsc{ImageNet} & \textsc{Places} & \textsc{Sun} & \\
        \hline
        Simple baseline & 77.0 & 70.4 & 61.1 (70.4) & 48.6 (56.5) & {\bf 61.3} & 4.59 \\
        Baseline with patches & 76.8 & 71.3 & 61.8 (71.8) & {\bf 49.2} (56.4) & 60.0 & 4.48 \\
        Mixture weights (MW) ($K=32$) & 76.6 & {\bf 71.9} & {\bf 63.1} (73.5) & 48.7 ({\bf 60.0}) & 58.8 & {\bf 4.29} \\
        MW with patches ($K=16$) & 74.9 & 70.3 & 60.9 (71.0) & 48.3 (56.2) & 58.8 & 4.6 \\
        MW with patches ($K=32$) & 74.9 & 70.3 & 62.7 (72.7) & 48.8 (57.8) & 58.0 & 4.4 \\
        MW with patches (diag. $K=16$) & 77.3 & 71.1 & 62.1 (71.6) & 48.9 (57.1) & 59.9 & 4.48 \\
        MW with patches (diag. $K=32$) & {\bf 77.7} & {\bf 71.7} & 62.3 (72.6) & {\bf 49.3} (57.9) & 60.0 & 4.4 \\
        \hline     
        \end{tabular}
        
        \label{tab:multi}
    \end{table*}

\paragraph{Mixture Weights initialization for a new task}
    As mentioned above, the choice of initial mixture weights $\Xi$ for a new task is crucial for efficient transfer learning, as our results in Table~\ref{tab:transfer} indicate that a wrong choice of initial mixture weights degrades the test accuracy on the new task after finetuning. One strategy is to choose the mixture weights $\Xi_\nu$ of the source domain $\nu$ that results in the best initial accuracy on the new data, but this might not be ideal since the new task may have similarities with \emph{multiple} tasks from the set of pretraining tasks. Therefore, we propose two simple solutions for the initialization of $\Xi$ that allows using multiple source mixture weights $\Xi_\nu$: accuracy-based and learned mixtures of mixture weights. In this part, we assume that all tasks share one set of labels to cancel the possible effects of finetuning the model head parameters, but our initialization techniques can be easily extended to an arbitrary supervised multitask case. In both approaches, we find the coefficients $c_\nu$ corresponding to each task $\nu$ from the set of pretraining tasks $V$ to compute the initial mixture weights $\Xi_{\nu}$ on the new task  $\nu^*$ as follows:
    \begin{equation}
        \Xi_{\nu^*} =\sum_{\nu \in V} c_{\nu} \Xi_\nu
        \label{eq:MofMW}
    \end{equation}
    The results in Table \ref{tab:MofMW} indicate that both initialization strategies.
    \textbf{Accuracy-based interpolation of $\Xi_\nu$.} Given the known accuracy $a_{\nu}$ on the new task $\nu^*$ for each task $\nu$ from the set of pretraining tasks, we compute the coefficients $c_\nu$ as $\frac{a_\nu}{\sum_{k \in V} a_k}$. Note that here we use the normalized accuracy values for computing the coefficients. \\
    \textbf{Learned interpolation of $\Xi_\nu$.}
    With the known ground truth labels on the new task, we can directly learn the coefficients $c_\nu$ that maximize the accuracy on the new task for the mixture weights constrained by the Equation~\ref{eq:MofMW}. 

    \begin{table}[ht]
        \centering
        \caption
        {
        Mixture weights initialization results for random initialization, accuracy-based interpolation and learned interpolation. The first part show the initial and accuracy after finetuning on the USPS dataset of a model trained on MNIST, Corrupted MNIST and SVHN; the second part shows the results on sketch domain of the model pretrained on other domains of the DomainNet dataset.
        }
        \begin{tabular}{lcccc}
        {\bf Dataset} & {\bf Random}  &  {\bf Acc-based} & {\bf Learned} & {\bf S/O} \\ 
        \hline
        Digits ft. & $93.76$ &  $95.50$ & $\bm{95.95}$ & $95.81$ \\
        Digits ft. & $96.05$ & $96.37$ & $\bm{96.61}$ & \\
        \hline
        DomainNet init. & $5.4$ & $\bm{15.08}$ & $13.57$ & $16.04 $ \\
        DomainNet ft. & $14.76$ & $\bm{16.44}$ & $15.90$ & $15.46$\\
        \hline
        \end{tabular}
        \label{tab:MofMW}
    \end{table}

\section{Unsupervised Domain Adaptation}
\label{apx:uda}

\subsection{Related work}
    Unsupervised domain adaptation (UDA) methods aim to learn a transformation of unlabeled target domain such that its distribution is close to the source domain distribution according to some metric of similarity. 
    There has been proposed a significant number of UDA methods. Some methods \cite{tzeng2017adversarial, long2017deep, tzeng2014deep, long2015learning} minimize the discrepancy between the source and target domain distributions by introducing an additional domain classifier; the target domain features are transformed so that the domain classifier cannot distinguish the domains accurately. Other works have explored feature moments matching \cite{zellinger2017central, peng2019moment} and second-order correlation \cite{sun2016return, peng2018synthetic} between source and target domain. Recent domain adaptation papers \cite{zhu2017unpaired, hoffman2018cycada, liu2017unsupervised} use GANs to minimize the pixel-level domain shift. 
    Some recent papers such as \cite{chang2019domain, wang2020fully} show the power of batch normalization for unsupervised domain adaptation. In fact, the concurrent work by \cite{wang2020fully} shows that fine-tuning of domain-specific batch normalization parameters alone gives promising results. We argue that our domain adaptation methods based on a compositional model described in Section~\ref{sec:uda} can be viewed as an extension of the domain-specific batch normalization since mixture weights affect the learnable parameters $\beta$ and $\gamma$ of batch normalization.

\subsection{Setup}

    \paragraph{Digits}
    For the experiment on digits datasets, we pretrained a classifier jointly on MNIST, Corrupted MNIST (shear), Corrupted MNIST (scale) and SVHN, and used USPS dataset as the target domain.
    The model contained 4 templates and 16 layers, used dropout with factor $0.3$, with a linearly decaying learning rate from $2 \times 10^{-3}$ to $2 \times 10^{-5}$; convolutional kernel weights regularization was set to $10^{-5}$.
    The model was trained for 2 epochs, at which point the accuracy over $97\%$ was reached for all source datasets except SVHN (over $92\%$).
    For the compositional models, we then copied the values of mixture weights of Corrupted MNIST (shear) to the target mixture weights since it gave the best source-only accuracy among source domains.
    We fine-tuned our model by minimizing the DA loss for 10 epochs. The total number of parameters in the classifier is $\approx 135000$. 

    \paragraph{DomainNet}
    For the experiment with DomainNet dataset, we selected 100 classes that occur most frequently in the dataset, and reshaped the images to $128\times 128$.
    We used infograph, clipart, painting, real and quickdraw as source domains, and sketch as target domain.
    For the baseline methods, we use an isometric model with 12 blocks that output tensors of shape $16\times 16\times 40$, and 12 templates and 16 blocks for the compositional models.
    We used the same learning rate, dropout and kernel regularization parameters as in the Digits experiment.
    We pretrained the classifier on the source domains for 20 epochs and fine-tuned them for another 20 epochs.
    For the compositional models, we copied the mixture weights from the painting domain to the target domain as it resulted in the best source-only accuracy.
    The classifier contained $\approx 340000$ parameters in total.

\subsection{Adversarial Discriminative Domain Adaptation}
\label{sec:adda_digits}

     After pretraining the model on source domain and copying the mixture weights, we started training the domain discriminator and target mixture weights simultaneously in an adversarial fashion.
     The discriminator was trained with Adam optimizer and a polynomially decaying learning rate from $2 \times 10^{-4}$ to $10^{-5}$.
     We compared the results of the compositional ADDA with those of the conventional ADDA model without weight sharing.
     The conventional baseline used a standard isometric model with the number of layers equal to number of templates in the compositional case and contained roughly same number of learnable parameters as our model.
     Our discriminator is a 3-layer CNN with a binary cross-entropy loss.

\subsection{Moment Matching Domain Adaptation}
    In this setup, we used an isometric model with 12 templates for the baseline method and a compositional model with 12 templates and 16 layers for compositional moment matching method. We pretrained our models for 20 epochs and finetuned them on target domain for 15 epochs. For the compositional model, we copied the mixture weights from the painting domain for before finetuning.
    For the moment matching, we used Adam optimizer with learning rate decaying from $10^{-4}$ to $10^{-5}$. Both models used dropout with drop probability $0.3$ and kernel regularization with weight $10^{-5}$.  

    \begin{table}[b]
        \centering
        \caption{
              Comparison of the transfer learning results for 3 datasets using: (a) a baseline model with a shared backbone (16 layers) and per-dataset heads; (b) mixture-weight (MW) models with 16 layers and either 16, or 32 templates.
        }
        \begin{tabular}{lccc}
        & \thead{\bf Baseline} & \thead{\bf MW Model \\ \bf (16 templates)} & \thead{\bf MW Model \\ \bf (32 templates)} \\
        \hline
        \textsc{EMNIST} & 93.5 & {\bf 93.7} & 93.1 \\
        \textsc{Aircraft} & 42.8 & {\bf 46.9} & 43.1 \\
        \textsc{CIFAR-10} & 93.5 & {\bf 93.7} & 93.1 \\
        \textsc{Stanford Dogs} & 63.2 & 64.0 & {\bf 67.5} \\
        \textsc{Cars196} & 57.5 & {\bf 63.4} & 63.1 \\
        \hline
        \end{tabular}
        \label{tab:transfer_ext}
    \end{table}

    \begin{table}[ht]
        \centering
        \caption{
        Supervised training test accuracy of a standard isometric model, compositional model with first half of the mixture weights shared across domains, and compositional model with separate mixture weights on Office-Caltech10 domains. The models were pretrained on the amazon, caltech and dslr domains simultaneously. 
        }
        \begin{tabular}{l c c c c}
        {\bf Domain} & {\bf Isometric} & {\bf Compositional} \\ 
        \hline
        amazon & $49.84$ & $\textbf{68.96}$\\
        caltech & $64.70$ & $\textbf{65.17}$\\
        dslr & $\textbf{64.54}$ & $55.51$\\
        \hline
        webcam & $\textbf{43.25}$ & $20.69$ \\
        \hline   
        \end{tabular}
        \label{tab:sup_adapt_digits}
    \end{table}

\subsection{Supervised Multi-Domain training results}
    \label{sec:sup_multi_da}
    \begin{table*}[ht]
        \centering
        \caption{
        Supervised training test accuracy of a standard isometric model, compositional model with first half of the mixture weights shared across domains, and compositional model with separate mixture weights on Digits domains. The models were pretrained on the MNIST, Corrupted MNIST (C-MNIST) shot noise, shear and scale, and SVHN domains simultaneously. 
        }
        \begin{tabular}{l c c c c}
        {\bf Domain} & {\bf Isometric} & {\bf Compositional half-shared} & {\bf Compositional} \\ 
        \hline
        MNIST & $99.14$ & $99.09$ &  $\textbf{99.29}$ \\
        C-MNIST (shot noise) & $ 97.63$ & $98.63$ & $\textbf{98.28}$ \\
        C-MNIST (shear) & $\textbf{98.79}$ & $98.69$ & $98.58$ \\
        C-MNIST (scale) & $98.18$ & $98.38$ & $\textbf{98.79}$\\
        SVHN & $54.73$ & $88.86$ & $\textbf{89.16}$ \\
        \hline
        USPS source-only & $95.77$ & $\textbf{96.37}$ & $96.05$ \\
        \hline   
        \end{tabular}
        \label{tab:sup_adapt_digits_2}
\end{table*}

\end{document}